\documentclass[sigconf]{acmart}

\usepackage{listings}
\usepackage{xcolor,colortbl}
\usepackage{tablefootnote}
\usepackage{fancyvrb}

\AtBeginDocument{%
  }

\copyrightyear{2024}
\acmYear{2024}
\setcopyright{acmlicensed}\acmConference[FDG 2024]{Proceedings of the 19th International Conference on the Foundations of Digital Games}{May 21--24, 2024}{Worcester, MA, USA}
\acmBooktitle{Proceedings of the 19th International Conference on the Foundations of Digital Games (FDG 2024), May 21--24, 2024, Worcester, MA, USA}
\acmDOI{10.1145/3649921.3649995}
\acmISBN{979-8-4007-0955-5/24/05}

\begin{document}

\title[You-Only-Randomize-Once]{You-Only-Randomize-Once: Shaping Statistical Properties in Constraint-based PCG}


 \author{Jediah Katz}
 \email{jediahkatz@gmail.com}
 \orcid{?}
 \affiliation{%
   \institution{Independent}
   \city{New York}
  \country{USA}
}

 \author{Bahar Bateni}
 \email{bbateni@ucsc.edu}
 \orcid{0000-0002-0701-0311}
 \affiliation{%
   \institution{University of California Santa Cruz}
   \city{Santa Cruz}
  \country{USA}
}

\author{Adam M. Smith}
\email{amsmith@ucsc.edu}
\orcid{0000-0002-4519-8423}
\affiliation{%
  \institution{University of California Santa Cruz}
  \city{Santa Cruz}
  \country{USA}
}

\begin{abstract}
In procedural content generation, modeling the generation task as a constraint satisfaction problem lets us define local and global constraints on the generated output. However, a generator’s perceived quality often involves statistics rather than just hard constraints. For example, we may desire that generated outputs use design elements with a similar distribution to that of reference designs. However, such statistical properties cannot be expressed directly as a hard constraint on the generation of any one output. In contrast, methods which do not use a general-purpose constraint solver, such as Gumin's implementation of the WaveFunctionCollapse (WFC) algorithm, can control output statistics but have limited constraint propagation ability and cannot express non-local constraints. In this paper, we introduce You-Only-Randomize-Once (YORO) pre-rolling, a method for crafting a decision variable ordering for a constraint solver that encodes desired statistics in a constraint-based generator. Using a solver-based WFC as an example, we show that this technique effectively controls the statistics of tile-grid outputs generated by several off-the-shelf SAT solvers, while still enforcing global constraints on the outputs.\footnote{Python code implementing our method can be accessed at \url{https://github.com/jediahkatz/you-only-randomize-once}.} Our approach is immediately applicable to WFC-like generation problems and it offers a conceptual starting point for controlling the design element statistics in other constraint-based generators.
\end{abstract}

       



\maketitle

\section{Introduction}

In procedural content generation (PCG), we often want our generated output to satisfy a set of hard constraints (such as reachability for certain key points in a generated level) \cite{pcgbook, sturgeon-org}.  Simultaneously, we would like the output to follow specific statistical properties (for instance, ensuring that the frequency of design elements used in the output is similar to the input \cite{guminWFC, context-sensitive}).

While constraint solvers provide a straightforward solution for handling hard constraints, they lack explicit mechanisms for integrating desired statistical properties. Even though many solvers offer \emph{optimization} criteria as a mechanism for expressing soft constraints, this does not work for statistical properties: We do not want an \textit{optimal} design, we want sampling of \textit{likely} designs.

In this paper, we introduce the \textbf{You-Only-Randomize-Once} (YORO) pre-rolling technique as a method to influence the output statistics of generic constraint solvers without the need to use a new solving algorithm or even modifying the existing solvers. By generating one batch of random numbers each time we are about to run the solver, we create a special decision variable ordering which results in outputs that respect the desired distribution. Part of what makes YORO remarkable is that it does not introduce any new source of randomness into the behavior of existing solvers.

To illustrate this idea in an easy-to-understand application familiar to the PCG research community, we apply it to generating 2D designs in the problem setting associated with the WaveFunctionCollapse (WFC) algorithm. Starting from an extremely simple example inspired by the Ising Model from statistical mechanics \cite{ising}, we scale up to a complex example involving replicating large-scale structures under path-based reachability constraints using a tileset from \emph{The Legend of Zelda}. Sampling several results from multiple black-box SAT solvers, we show that YORO delivers on the promise of statistical control through decision variable order manipulation.


\section{Related Work}
\label{sec:related-work}
YORO bridges three distinct areas of research: \textbf{Procedural Content Generation}, \textbf{Constraint Solving}, and \textbf{Statistical Sampling}. Curiously, while researchers have explored the intersection of every pair of these topics, the trio is rarely combined.

The connections between PCG and constraint solving have been explored by Smith and Mateas's applications of answer-set programming (ASP) to PCG \cite{asp4pcg}. By capturing the target design space as a declarative definition, taking the form of an answer-set program, they aim to directly write down the properties that each generated output must exhibit. Similarly, Cooper's Sturgeon describes its output artifacts (i.e. tile-based game levels) by specifying a set of constraints \cite{sturgeon-org}. Further, Sturgeon is able to incorporate not only pattern rules extracted from a set of examples, but also complex constraints such as path-based reachability of certain key points in the level, resulting in the generation of guaranteed-playable game levels. Additionally, Sturgeon also specifies a set of frequency rules with the goal of applying the desired statistical properties. These rules constrain the output tile count over certain tags and regions to be within a margin of those in the example data. One important distinction between Sturgeon's approach to enforcing statistical properties and our proposed method, YORO, is that YORO ensures these properties exist on a large enough population of outputs as opposed to every single acceptable artifact in the design space. As a result, the expressive range of the PCG system is unaffected by the inclusion of the desired statistical properties. In other words, the number of possible artifacts in the design space is the same since the definition of this space has not changed (only the distribution).

Later, Cooper expanded on the idea of incorporating more complex constraints into the definition of the design space by introducing Sturgeon-MKIII \cite{sturgeon-MKIII}. By defining the game mechanics as rewrite rules, Sturgeon-MKIII simultaneously generates the level and a playthrough of it, thus ensuring the playability of the generated level. Looking forward, we would like to be able to express statistical knowledge as well.

In the separate context of placing objects in indoor levels, Horswill and Foged propose path constraints as a way to define a wide variety of design constraints \cite{horswill-path-constraints}. These constraints range from lock-and-key problems to guaranteeing the survivability of the level by a careful placement of monsters and health packs. This is done by first defining path functions which summarize an attribute over some path on a graph. These attributes can be, for example, the expected health loss or gain in each node. The system is able to then define specific constraints on these functions, which are considered during the constraint solving process. Furthermore, the fast calculating of these functions made possible through dynamic programming allows for even real-time applications.

Connecting constraint solving and statistical sampling, probabilistic logic programming systems (e.g. Markov Logic \cite{markov-logic} or Probabilistic Soft Logic \cite{bach2017hinge}) offer modeling languages reminiscent of ASP but with inference engines capable of sampling from precisely specified distributions and even adapting the definition of those distributions to fit example data. The related literature on nearly-uniform samplers \cite{golia2021designing} and weighted model counting \cite{chakraborty2014distribution} also connect these worlds. These advanced systems and techniques may serve as the foundation for content generation systems in a distant future, but the available literature currently offers no guides for how the PCG practitioner should attempt to use them.

Finally, statistical sampling is connected to procedural content generation most obviously via the paradigm of Procedural Content Generation via Machine Learning (PCGML) \cite{pcgmlbook}. Despite the importance of hard design constraints like reachability, PCGML systems often try to learn these properties from example designs rather than allowing users to directly specify what they want (potentially requiring them to need to learn a formal specification language first). It is not obvious how to provide current PCGML systems with additional symbolically-encoded background knowledge.

Specifically in the context of the WaveFunctionCollapse algorithm, seeking certain statistical properties in the output can improve the generated results. When introducing WFC, Gumin highlighted one of the main goals of WFC as similarity between the distribution of patterns in input and a sufficiently large number of outputs \cite{guminWFC}. To achieve this, his algorithm used randomization during the constraint solving process to heuristically make local choices following the marginal distribution seen in the example input designs. By contrast, our method concentrates all of the randomization in a preprocessing step that runs before an existing solver runs to produce an output design. By factoring the statistical concerns out of the solver's search process, we gain the ability to drop in alternate constraint solvers.

Recently, Bateni et al.\ expand on Gumin's desire for \emph{resemblance} by introducing Context-Sensitive WFC \cite{context-sensitive}. By modeling the distribution of tiles or patterns conditioned on their surrounding context, they demonstrate the significance of leveraging statistical properties in both the quality and expressive range of the output. Importantly, they show that Gumin's tile-level heuristic was insufficient to achieve its original goal. Reproducing neighborhood-level statistics required using a neighborhood-level statistical model. While Bateni's method yields WFC-like results with greatly improved resemblance, it inherited WFC's limitations: it could not incorporate global constraints such as reachability. By using YORO we aim to keep the advantages of solver-based methods in employing global constraints while also gaining some control on the statistical characteristics of the output.

\section{Technical Background}

\subsection{Satisfiability Solvers}
Satisfiability (SAT) solvers are programs that solve the classic NP-complete Boolean satisfiability problem. They are widely used to solve a number of practical problems whose constraints can be represented as a Boolean formula. SAT solvers accept as input a description of a Boolean formula in conjunctive normal form (CNF), i.e. an ``AND of ORs,'' and output either a satisfying assignment or the message \texttt{UNSAT} to indicate that no satisfying assignment exists.

In a typical SAT solver, Boolean variables $x_1, x_2, \dots, x_n$ are represented programmatically as integers $1, 2, \dots, n$, and a Boolean literal is a variable in positive form (e.g., $x_7$ is $7$) or negative form (e.g., $\neg x_7$ is $-7$). SAT problems are typically represented in conjunctive normal form (CNF): one big conjunction (AND) of many small disjunctions (OR), each involving one or more positive or negative literals. An input CNF formula is represented as a list of clauses, which themselves are lists of literals. For example, an input to a SAT solver might be the formula $(x_1 \vee x_2) \wedge (\neg x_1 \vee \neg x_2)$, represented as $\texttt{[[1, 2], [-1, -2]]}$. An output for this formula might be the satisfying assignment $\{x_1 = \text{True}$; $x_2 = \text{False}\}$, represented as $\texttt{[1, -2]}$. Another possible satisfying assignment is \texttt{[-1, 2]}. By default, most SAT solvers terminate after yielding the first satisfying solution they find, but most can be configured to continue enumerating additional solutions. In this paper, we will focus on influencing just the very first solution output by a solver.

\subsection{Decision Variable Ordering} \label{sec:variable-ordering}
When representing an abstract constraint satisfaction problem as a CNF formula, a \textit{variable ordering} refers to a labeling of semantically-named Boolean variables with integers from $1$ through $n$. Although a Boolean formula has the same set of solutions regardless of how its variables are ordered, the choice of ordering can impact which of those solutions the solver will output first.

During the execution of a typical SAT solver, the solver attempts to incrementally build a satisfying assignment by \emph{selecting} unassigned variables one at a time and then \emph{deciding} a value for them (True or False). In solvers employing constraint propagation methods, many variables are assigned values deduced from the value of previously assigned variables, and the solver's variable selection mechanism is only invoked when there is no other work that must be done first. With or without constraint propagation, solvers can encounter situations where there are no longer any values available to be assigned to a variable (i.e., the solver's previous choices have been revealed to be contradictory). To resolve this contradiction, many solvers backtrack (undoing one or more recent choices) before trying to make an alternate choice.

Which unassigned variable should a solver \emph{select} next? Many heuristics have been developed to determine the order in which a solver chooses variables for the next decision step. SAT solvers may use \textit{static} orderings, which are fixed at the beginning of solving, or \textit{dynamic} orderings, which change over the course of solving. Jeroslow-Wang is an example of a common static heuristic, in which variables are ordered based on the frequency of their appearance in the input formula ~\cite{jeroslow-wang}. VSIDS is an example of a common dynamic heuristic, in which variables move up in the ordering if they cause a contradiction to occur ~\cite{chaff}. However, when the order of decisions is not influenced by heuristics, SAT solvers will typically decide variables in ascending order of the variable ordering. Many configurable solvers even provide the ability to fully disable heuristics, falling back to a selection order based on the numerical representation used in the CNF formula. In YORO, we manipulate this ordering so that a solver's fallback strategy is to let the target statistics guide the selection order.

\subsection{WFC as a Boolean CSP}\label{sec:wfc-sat-encoding}
The grid-based WaveFunctionCollapse algorithm is usually seen as having two phases: input analysis, in which the input grid is processed to extract the set of tiles and the allowed adjacencies between tiles, and then grid generation, in which tiles are assigned to a new grid while respecting the allowed adjacencies. Here, we will show how the generation phase of WFC can be implemented using any SAT solver. Recall that the goal of WFC is to find an assignment of tiles to grid cells such that tiles are only adjacent in the output if they were seen to be adjacent in the input example.

Assume we have already extracted the set of tiles $T$ and a set of adjacency lists \texttt{right[t]} and \texttt{below[t]}, which represent the set of tiles that may be placed immediately to the right and below a tile $t$, respectively. Let $N \times M$ be the desired dimensions of our output grid. For simplicity we assume that grids have periodic boundary conditions, i.e. row $N+1$ and column $M+1$ are understood to refer to row $1$ and column $1$, respectively.

We first define the following Boolean variables:

\vspace{3pt}
\begin{center}
\texttt{assign} ($x, y, t$) $\iff$ the cell at position ({$x, y$}) is assigned tile t
\end{center}
\vspace{3pt}

These Boolean variables can be arbitrarily labeled from $1$ to $N \cdot M \cdot |T|$, for example with the mapping
\vspace{3pt}
\begin{center}
{\texttt assign}($x, y, t_i$) $\mapsto$ ($x \cdot M + y$) $\cdot$ |T| + i
\end{center}
\vspace{3pt}

Next, we construct Boolean clauses to represent the constraints of our WFC setting. Assume we have a function \texttt{add\_clause()} which adds a clause to the growing CNF formula. First, we have the constraint that each cell of the grid must be assigned \emph{at least} one tile.

\begin{lstlisting}[mathescape=true]
for each cell position $(x, y)$:
    add_clause$\left( \displaystyle\bigvee_{\texttt{tile } t} \texttt{assign(}x, y, t\texttt{)}\right)$
\end{lstlisting}

Next, we have the constraint that each cell of the grid must be assigned at \emph{most one} tile. We can equivalently state this as, ``for each pair of distinct tiles, they cannot both be assigned to one cell.''

\begin{lstlisting}[mathescape=true]
for each cell position $(x, y)$:
    for each pair of distinct tiles ${t_1, t_2}$:
        add_clause( $\neg$assign($x, y, t_1$) $\vee$ $\neg$assign($x, y, t_2$) )
\end{lstlisting}

Finally, we have the constraint that assigned tiles must respect allowed adjacencies. We can equivalently state this as ``if a cell is assigned tile $t$, then its adjacent cell must be assigned to a tile which is an allowed adjacency for $t$.'' On a two-dimensional grid, we must enforce constraints for both horizontal and vertical adjacencies.

\begin{lstlisting}[mathescape=true]
for each cell position $(x, y)$:
    for each tile $t$:
        a = assign(x, y, t)
        add_clause$\left( \neg{}a \vee \displaystyle\bigvee_{t_r \texttt{ in right}[t]} \texttt{assign(}x + 1, y, t_r\texttt{)}\right)$
        add_clause$\left( \neg{}a \vee \displaystyle\bigvee_{t_b \texttt{ in below}[t]} \texttt{assign(}x, y + 1, t_b\texttt{)}\right)$
\end{lstlisting}

At this point, we have a CNF formula that we can give to a SAT solver, and which we assume outputs a satisfying assignment. To decode that assignment into an output grid, we can simply identify which Boolean variables \texttt{assign($x, y, t$)} are True in the satisfying assignment, and assign tile $t$ to the cell at $(x, y)$ in the output grid.

There are many other ways to reduce the WFC-inspired grid generation problem to SAT, but we have chosen a clean and simple one here to illustrate how to apply the YORO technique.

\subsection{Gumbel-max Trick} \label{sec:gumbel}
Before we introduce YORO, we should note where others have drawn a theoretical connection between the procedure of sampling a distribution without replacement and generating a single stochastic ordering of the items to be sampled. Recall that in Gumin's WFC, a \emph{tile frequency heuristic} is used to sample the next tile assignment to try during search from the pool of tiles remaining in a cell based on some distribution. We want to mimic within-search randomization (something that might require significant engineering effort to add to an existing constraint solver) by way of preparing a clever static ordering.

The Gumbel-max trick \cite{gumbel,huijben2022review} is a widely applied method of sampling from a categorical distribution with un-normalized weights $w_i$ for each class $i \in [1..k]$. The Gumbel-max trick separates the distribution into a constant term, which is defined by the log-weights of each class, and an independent Gumbel noise term. The Gumbel noise term is a random sample $G_i$ from the $\text{Gumbel}(0, 1)$ distribution, which can be conveniently and accurately approximated by $G_i \sim -\log(-\log(\text{Uniform}(0, 1))$.The following is equivalent to choosing a category $y$ by a weighted random sample:
$$ y = \underset{i \in [1..k]}{\text{argmax}}\left(\log(w_i) + G_i\right) $$
An extension to the Gumbel-max trick allows for repeated sampling without replacement to create a permutation \cite{gumbel-sequence}. If we arrange the classes $i$ in decreasing order of their values $\log(w_i) + G_i$, this is equivalent to sampling without replacement $k$ times: first from the full set of categories, then from the remaining $k-1$ categories based on their collective weights, and so on.

In a moment, we will show how the Gumbel-max trick can be employed to sort our decision variables \texttt{assign($x, y, t_i$)} into an order that approximates within-search sampling from the desired tile distribution.

\section{Our Technique: YORO Design Pre-rolling}

\begin{figure*}[h]
  \includegraphics[width=1.7\columnwidth]{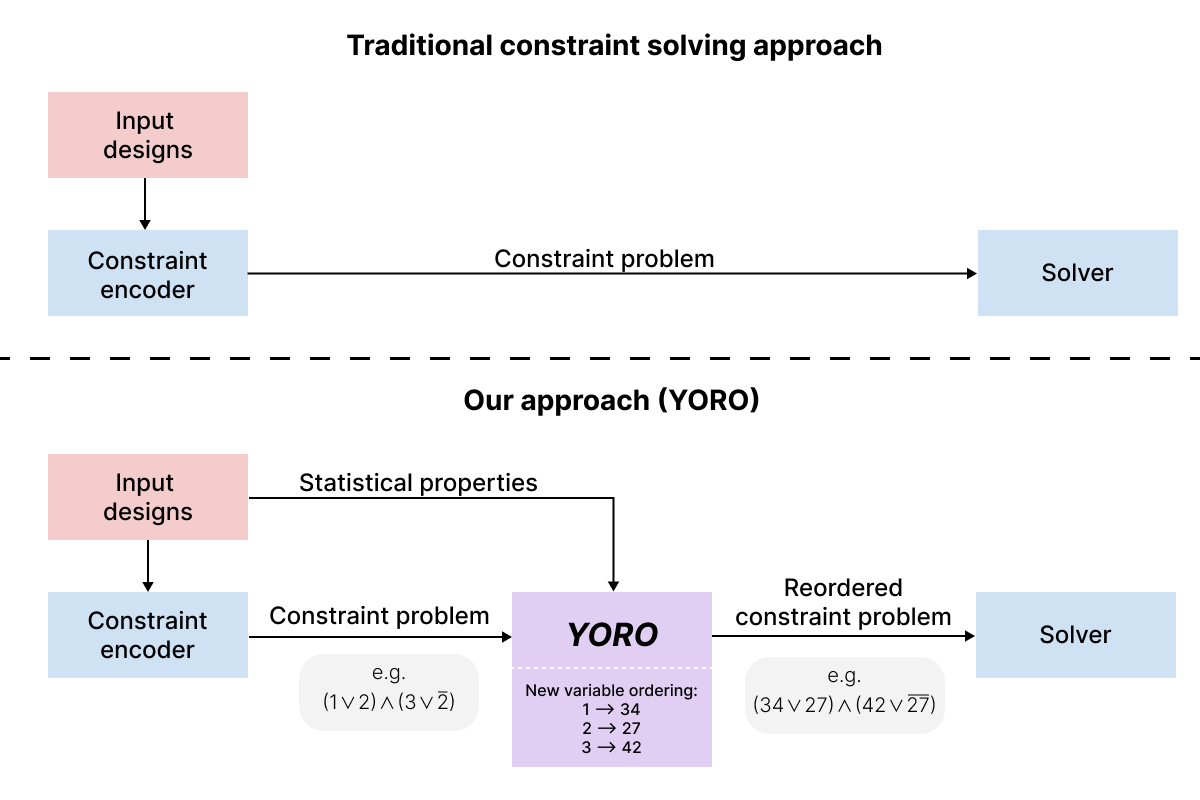}
  \caption{In the traditional constraint solving approach to procedural content generation, it is assumed that all relevant design concerns can be represented as constraints that must be satisfied by each output design in isolation. In other words, the solver’s job is to produce outputs that are guaranteed to be free from easily detectable flaws. This approach cannot capture population-level design considerations such as that design elements should be used with some typical frequency across the space of outputs even though the allowed frequency of those design elements is largely unrestricted within a single output. Our technique, YORO, transforms a constraint program so that the first solution output by the solver after each randomization is more likely to represent the target statistics.}
  \label{fig:approach-comparison}
  \Description{This is the description.}
\end{figure*}

Figure~\ref{fig:approach-comparison} compares the YORO approach with a traditional approach to constraint solving for PCG applications (e.g.\ the \emph{design-space modeling} paradigm sketched by Smith and Mateas \cite{asp4pcg}). With YORO, we preprocess the low-level definition of a constraint problem before the solver gets to look at it. This manipulation is intended to shape the statistical properties evident within and across the collection of first-solutions output by the solver after each randomization.

As mentioned in Section~\ref{sec:variable-ordering}, when a SAT solver has no other heuristics to apply, it will typically fall back to the variable ordering implied by the problem specification to break ties. In this way, by curating the default order, we can bake our statistical desires into the solver’s tie-breaking behavior without needing to modify the solver at all. To achieve this, we craft a variable ordering using pre-rolled Gumbel noise for each cell that samples from the desired distribution of design elements.

Suppose we want to control the tile frequency statistics in the WFC setting, such that outputs follow a distribution which gives tile $t$ a probability of $P[t]$. Assume we have defined Boolean variables \texttt{assign}$(x, y, t)$ as described in Section~\ref{sec:wfc-sat-encoding}, for which we now must craft a variable ordering. We can do so with the following pseudocode, which defines a Python-style sorting key function, such that variables will be sorted based on their key value:

\begin{lstlisting}[mathescape=true]
variables = [assign($x, y, t$) for pos ($x, y$), for tile $t$]
variables.sort(key=sorting_key) 

function sorting_key( assign($x, y, t$) ):
    cell_pos_rowmajor = ($y, x$)
    gumbel_noise = -log(-log(random(0, 1)))
    tile_score = log($P[t]$) + gumbel_noise
    return (cell_pos_rowmajor, -tile_score)
\end{lstlisting}


First, we choose an arbitrary, fixed ordering for the cell positions (e.g., row-major order from the top-left).\footnote{This ordering for enumerating the cells on a grid corresponds to the \emph{lexicographic} selection heuristic that Karth \cite{karth} found to perform similarly to Gumin's \emph{entropy} heuristic. The effect of each of these is that the solver will make its next selection very close to where it had made previous selections, which is also often a location where there are relatively few remaining options for the tiles that may be placed in a cell.} This determines the order in which the solver will choose which cells to assign a tile, and depending on the constraints of the problem, this choice may lead to bias. We use the cell position as the primary sorting key.

Next, we use the Gumbel-max trick to sample a \texttt{tile\_score} based on the probability for each tile. This score is used as the secondary (tie-breaking) sorting key, which determines the ordering of tiles for all variables with the same cell position. As described in Section~\ref{sec:gumbel}, sorting the set of tiles by this \texttt{tile\_score} is equivalent to sampling from the set without replacement repeatedly. Note that the \texttt{tile\_score} is negated so that when the solver works through the variables in order, it often tries the more likely options first.\footnote{It is important to note that the solver should not simply decide variables in \emph{order of decreasing likelihood}. This would result in the solver deterministically making the same decisions each time it is re-run. High-likelihood choices should go earlier in the order, but only with a controlled level of randomness in just how early they go (given by the Gumbel noise term).}

\begin{figure*}[h]
  \includegraphics[width=\textwidth]{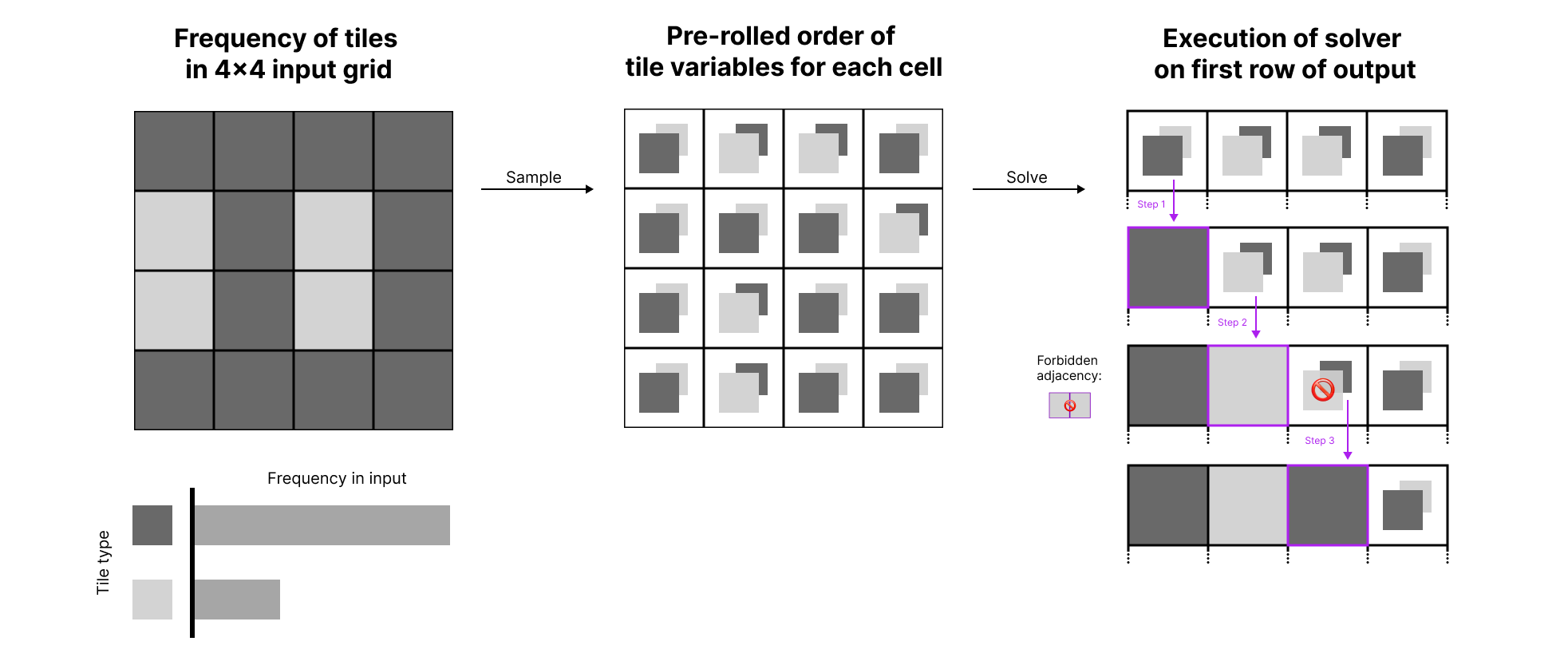}
  \caption{YORO applied to a toy $4 \times 4$ grid generation task where black (dark) tiles are expected to be seen more often than white (light) tiles. Left: The input design establishes the vocabulary of allowed tiles, allowed tile adjacencies in different directions, and the target distribution for using each tile in new designs. Middle: Drawing from the target distribution at each cell, we come up with a fixed ordering of the tiles within each cell. In this sampled ordering, white will be explored first in 5 out of 16 cells (note that white was observed to be used 4 out of 16 cells in the input grid). Right: Four steps of constraint solving search in which the next available un-assigned variable is chosen and the implications of that choice are propagated out to nearby cells. After these first four steps, white tiles will have been placed in just 1 out of 4 of the top-row cells, matching the target distribution from the input image.}
  \label{fig:pre-roll-example}
  \Description{This is the description.}
\end{figure*}

Figure~\ref{fig:pre-roll-example} narrates the details of using the YORO technique to generate a $4 \times 4$ output that targets the tile frequency statistics of a simple, black-and-white input grid of the same shape. The process of encoding WFC into a CNF formula is not pictured here; instead, this diagram simply demonstrates sampling a variable ordering with YORO and shows how a solver assigns tiles based on the sampled ordering and the adjacency constraints.

In the first section, we divide the input into tiles and compute the frequency of each tile. The second section represents the construction of a variable ordering using the YORO method. Each cell in the diagram pictures the two Boolean variables that correspond to assigning a black or white tile to that position in the output, respectively. For each cell position, a new ordering is sampled with a 75\% probability of choosing black first and a 25\% probability of choosing white first. Note that only 4 of 16 tiles are white in the input, but the YORO sub-orderings (i.e., the orderings of variables for the same cell) for 5 of 16 cells place white first. Such deviations are common and expected due to the random nature of the process.

In the third section, we represent the inner execution of a SAT solver as it assigns the first three tiles. We assume the SAT solver decides variables in ascending order of the provided ordering. We also assume our variable ordering is in row-major order of cell positions, so the first three cells decided are at $(0, 0)$, $(1, 0)$, and $(2, 0)$. The very first Boolean variable in the ordering is \texttt{assign}$(0, 0, \texttt{B})$, so the solver assigns the first cell to black. The second variable in the ordering is \texttt{assign}$(0, 0, \texttt{W})$, but since the first cell is already black and our constraints enforce that each cell can be assigned at most one tile, the solver infers that this variable must be false. The next variable in the ordering is \texttt{assign}$(1, 0, \texttt{W})$, so through a similar process the solver assigns the second cell to white. Finally, the third sub-ordering begins with \texttt{assign}$(2, 0, \texttt{W})$, so the solver will try to assign the third cell to white. However, since two white tiles never occur horizontally adjacent in the input, this assignment will violate the adjacency constraints. The solver will then detect a contradiction and be forced to backtrack, assigning the third cell to black instead.\footnote{Actually, most solvers will have already inferred this without needing to backtrack.} In effect, our fixed (but randomly generated) variable ordering has allowed the solver to draw an appropriate sample among the options remaining after constraint propagation.

\section{Experiments}

\subsection{Tile-level Pre-rolling}

\begin{figure*}[h]
  \includegraphics[width=\textwidth]{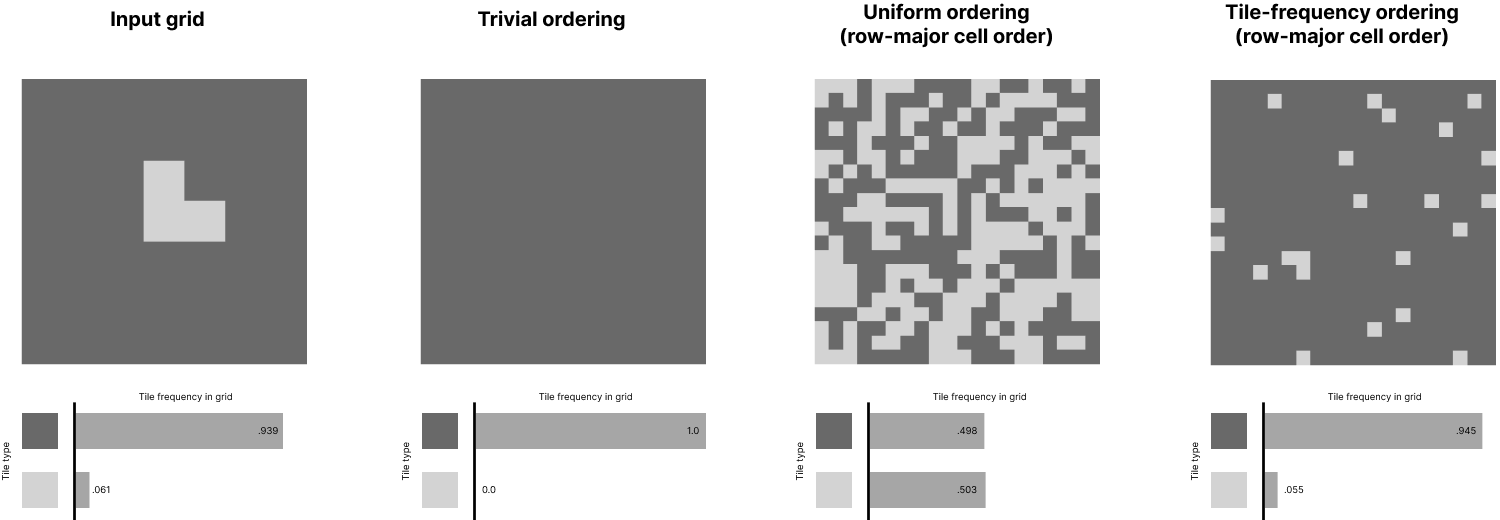}
  \caption{Exploring variable orderings for a $20 \times 20$ toy example where any tile can be placed next to any tile, but black tiles should be much more common than light tiles. Trivial: Without manipulating the variable ordering, the solver produces a valid-but-undesirable image composed of only black tiles. Uniform: By manipulating the variable ordering, we can inject diversity into the solver's outputs, but the statistics are far from the target. Tile-frequency: Applying the right kind of randomization to the variable ordering, we get outputs where close adherence to the target statistics is obvious from even a single output sample.}
  \label{fig:black-and-white-outputs}
  \Description{This is the description.}
\end{figure*}

In our first experiment, we sought to craft a simple and easily understood example to demonstrate the impact of the YORO method on the generated output statistics without any influence by constraints. To that aim, we defined a $7 \times 7$ example grid with only two colors: black and white (displayed as dark and light gray in this paper for contrast). Nearly all the cells are colored black except for three, which are arranged in an L-shape pattern. Crucially, note that the adjacencies in this input grid are such that all arrangements of black and white tiles are allowed; that is, there are no adjacency constraints in this example. Therefore, we expect that generated outputs based on this grid should exhibit tile frequency statistics based purely on the order that tiles are assigned in. Figure~\ref{fig:black-and-white-outputs} illustrates this scenario and previews impacts of variable orderings.

This scenario where each location can take on just one of two states is closely related to the Ising model from statistical mechanics \cite{ising} where the $+1$ or $-1$ spin configuration of atoms in an idealized two-dimensional grid is analyzed. Two physical effects are typically captured with this style of model: a global tendency for certain spin states to be seen more often than other (e.g. $+1$ might be more common than $-1$ because of the application of an external magnetic field), and a local tendency for the state of one atom to agree or disagree with the state of neighbors (modeling the strength of local atomic interactions). Our demonstration of YORO in this section explores only how to model the global tendency to use tiles with a target distribution, and a later section will explore how to capture the statistics of local interactions.

Continuing, we generated several $20 \times 20$ output grids using the single-tile-based WFC encoding described in Section~\ref{sec:wfc-sat-encoding}, solved with Google's OR-Tools solver. We compare three methods of crafting a variable ordering for the solver. First, we provide a \textbf{trivial ordering}, in which variables are sorted lexicographically by their cell position and tile index. This ordering is representative of the ordering that might be output by the grounder of an answer-set programming system that generates low-level variables and constraints by expanding compact formulae in the system's high-level modeling language. Next, we use the YORO technique to construct another variable ordering (\textbf{uniform ordering}) in which variables are arranged primarily in row-major order of their cell position, and variables with the same cell-position are secondarily arranged in a uniformly-random order. This ordering should show crude statistical control over outputs (yielding noticeable diversity within and across solver outputs) without yet aligning those statistics with those of the target. Finally, we use the YORO technique to craft a variable ordering in which variables are arranged primarily in row-major order of their cell position, and secondarily in random order sampled based on \emph{tile frequency}. This order is intended to yield results with statistics approximating those of the target.

The output generated with the trivial ordering is uniformly black, with no white tiles at all. Since black happens before white in our lexicographic order, the solver attempts to assign each cell to black first. There are no adjacency constraints to forbid every tile from being assigned black, resulting in the final output.

The output generated with the uniformly random tile-level ordering reflects the random sampling result from a rudimentary application of variable order manipulation. For each cell, the solver will select its corresponding black and white variables in the provided order. The tile frequency statistics for this output are split almost exactly equally between black and white tiles. Additionally, the black and white tiles appear to be distributed randomly throughout the grid, rather than being organized into structured groups. Absent any consideration for target statistics, this rudimentary result shows one way to inject diversity into the outputs of generators based on black-box constraint solvers.

Finally, the output generalized with the tile-frequency based random ordering illustrates the ability of YORO to closely match the distribution of the input grid almost perfectly. In this case, the frequency of black and white tiles is nearly identical between the input and output grids with an error of less than 1\%. It is worth noting that this output still does not fully \emph{resemble} the input grid at a pattern level, as it contains only one L-shape pattern. This is a result of single-tile formulation of WFC used for this experiment, which only accounts for the raw frequency of tiles and their valid adjacencies and does not consider the distribution of multi-tile patterns. 

Our results here corroborate those of Bateni et al.~\cite{context-sensitive}; even a highly effective tile-frequency heuristic can yield results that fail to resemble input designs even though the WFC algorithm is otherwise so closely focused on reasoning about tile adjacencies. In terms of the Ising model, we are missing a statistical model of the interactions between adjacent sites on the grid that will help us break ties when there are many tiles still available for selection according to the hard constraints.

\begin{figure*}[h]
  \includegraphics[width=\textwidth]{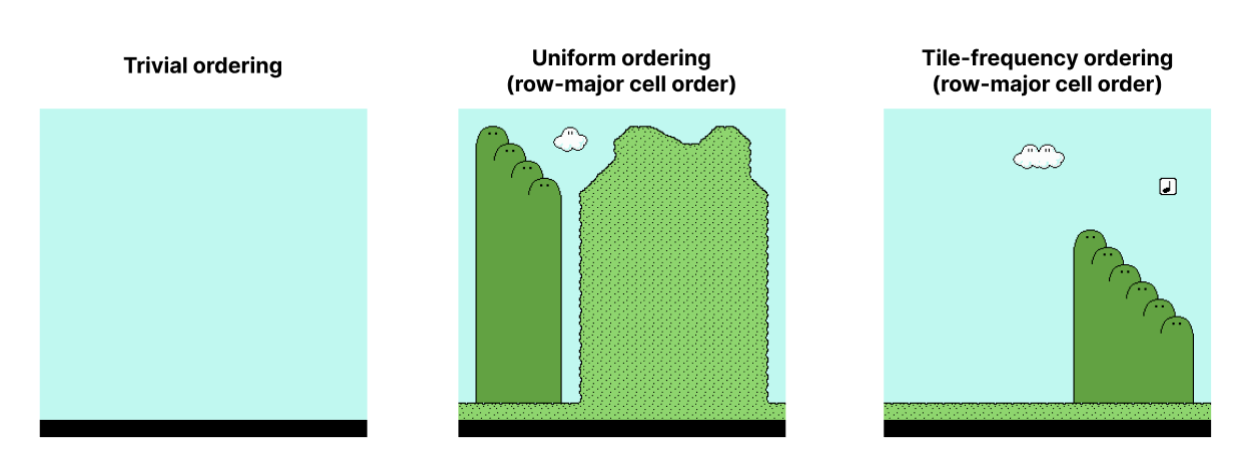}
  \caption{Exploring variable orderings in the \emph{Mario} domain. Compare with Figure~\ref{fig:black-and-white-outputs}. In this example, we are not yet modeling neighborhood-level statistics or attempting to enforce any interesting global constraints. Nevertheless, the YORO technique is able to immediately improve upon the results that would be seen without any decision variable ordering manipulation.}
  \label{fig:mario-outputs}
  \Description{This is the description.}
\end{figure*}

Surprisingly, just enforcing tile-level statistics is sufficient to get interesting results for game-related content generation tasks. In Figure~\ref{fig:mario-outputs}, we apply the same process to the above-ground section of the World 1-2 map from \emph{Super Mario Bros 3.} for the NES \cite{mario-nes}. The only modification used in this scenario to disable periodic boundary conditions for the output grids for aesthetic reasons.

\subsection{Neighborhood-level Pre-rolling with Global Constraints}

\begin{figure*}[h]
  \includegraphics[width=\textwidth]{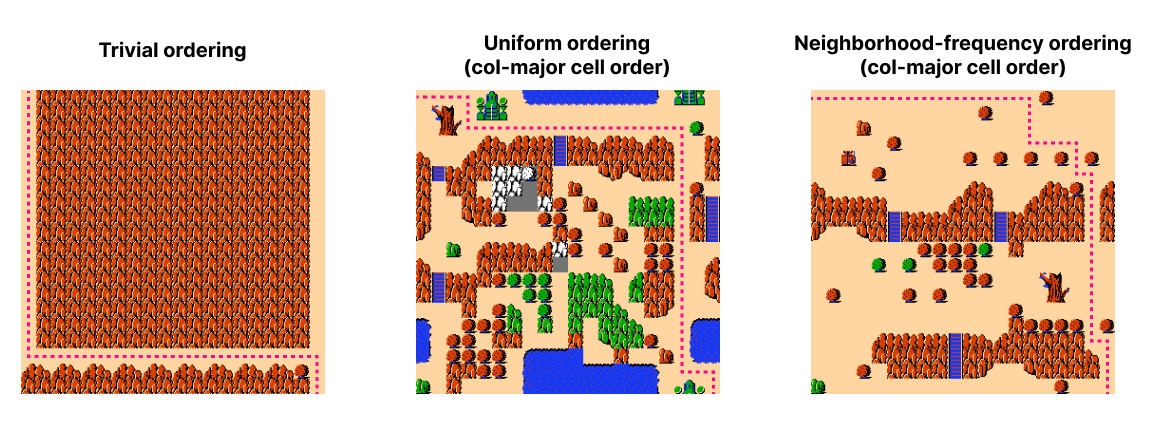}
  \caption{Exploring variable orderings for the \emph{Zelda} domain. Here, we approximate Bateni's context-sensitive WFC method \cite{context-sensitive} by sampling design choices at the level of whole neighborhoods. Further, we enforce an example global constraint: there must be a beige dirt-tile path (marked in pink) from the top-left to the bottom-right that only moves in downward or rightward steps.}
  \label{fig:zelda-outputs}
  \Description{This is the description.}
\end{figure*}

In the second experiment, we attempted to demonstrate the efficacy of using the YORO method in a realistic setting to control output statistics even while enforcing an interesting global constraint. For our input grid, we used the overworld map from \emph{The Legend of Zelda} for the NES, pictured in Figure \ref{fig:zelda-outputs}, which consists of 90 unique tiles of $16 \times 16$ pixels \cite{zelda-nes}.

To achieve outputs with closer resemblance to the input grid, we used a more complex SAT formulation of WFC based on Bateni’s \emph{context-sensitive} decision heuristic \cite{context-sensitive}. The context-sensitive decision heuristic is a method to determine which tile a WFC generator should choose when assigning a cell. Rather than sampling a tile based on individual tile frequency, it samples based on the joint frequency of the tile and its four-tile neighborhood (i.e., the adjacent north, east, south, and west tiles) in the input image, accounting for cells that haven’t been assigned yet. When the neighborhood for the current cell to be assigned does not exist at all in the input, the heuristic falls back to sampling based on individual tile frequency.

It may not be possible to directly reproduce Bateni’s context-sensitive heuristic within SAT-based WFC implementation without a custom dynamic heuristic for the SAT solver. However, we can approximate it by introducing a new set of neighborhood-assignment variables. We define the following variables for each tile $t$ and neighborhood $(t, t_n, t_e, t_s, t_w)$:
\begin{equation*}
\texttt{assign}(x, y, t, t_n, t_e, t_s, t_w) \iff 
\begin{aligned}
    &(x, y) \text{ is assigned $t$ and its} \\
    &\text{neighboring tiles are assigned } \\
    &t_n, t_e, t_s, t_w \text{ respectively}
\end{aligned}
\end{equation*}

These Boolean decision variables are defined in addition to the individual-tile-assignment variables (\texttt{assign($x$,$y$,$t$)}). Then, we use YORO to craft a variable ordering such that for each cell, its sub-ordering consists first of the neighborhood-assignment variables and then the tile-assignment variables. The neighborhood-assignment variables are in a randomly sampled order based on neighborhood frequency in the input, and the tile-assignment variables are similarly arranged based on a tile frequency sampling. With this formulation, the solver will assign entire neighborhoods at a time, and if no neighborhoods present in the input are possible then the solver will fall back to assigning individual tiles based on tile frequency.

For this experiment, also we add a simple global constraint to all generated outputs: there must be a path of only dirt tiles from $(0, 0)$ to $(N-1, M-1)$ that moves only rightwards and down (henceforth a “good dirt path”). We can represent this constraint in SAT by adding new variables defined as follows:
$$\texttt{reachable}(x, y) \iff \text{a good dirt path connects $(0, 0)$ and $(x, y)$}$$ 

We then add recursive constraints to enforce that $(x, y)$ is reachable iff $(x, y)$ is a dirt tile and $(x-1, y)$ or $(x, y-1)$ is reachable, handling the base cases where $x = 0$ or $y = 0$ separately. Also note that we place all the \texttt{reachable}$(x, y)$ variables at the end of the variable ordering, since we do not want the solver to make decisions based on them.

We once again compare three methods of crafting a variable ordering: a trivial variable ordering; a uniform ordering, in which neighborhoods are permuted randomly in the YORO sub-orderings for each cell; and a neighborhood-frequency YORO ordering as described above. Initial results are shown in Figure~\ref{fig:zelda-outputs}.

The trivial ordering places mostly rock tiles, which happen to have the lexicographically smallest index of $0$. However, it is forced to change some of these to dirt tiles in order to satisfy the global constraint with a simple good dirt path.\footnote{The trivial ordering produces identical results across two runs, since there's no randomness involved.} Note that the final column is also assigned dirt tiles; since the solver prefers to assign entire neighborhoods at once, and the grid has periodic boundary conditions, the west neighbor of the first column wraps east.

In the outputs generated with the uniform and neighborhood-frequency YORO orderings, we begin to see more structured groups of tiles as a result of the neighborhood-based formulation. However, the uniform output contains a wider variety of tiles, which are positioned more sporadically, while the neighborhood-frequency output is more sparse and contains a more homogenous sample of tiles that appear frequently in the input. That its, the outputs reproduce more of the large-scale structures seen in the input design that consist of multiple neighborhoods. Also note that the global constraint is satisfied in both outputs, but in a rather lazy manner, where the dirt path avoids turning until near the end of the row. Future work might try to improve the aesthetics of this path using the YORO method, by having the reachability variables participate meaningfully in the variable ordering. 

\section{Adapting YORO for Different Solvers}

\begin{figure*}[h]
  \includegraphics[width=\textwidth]{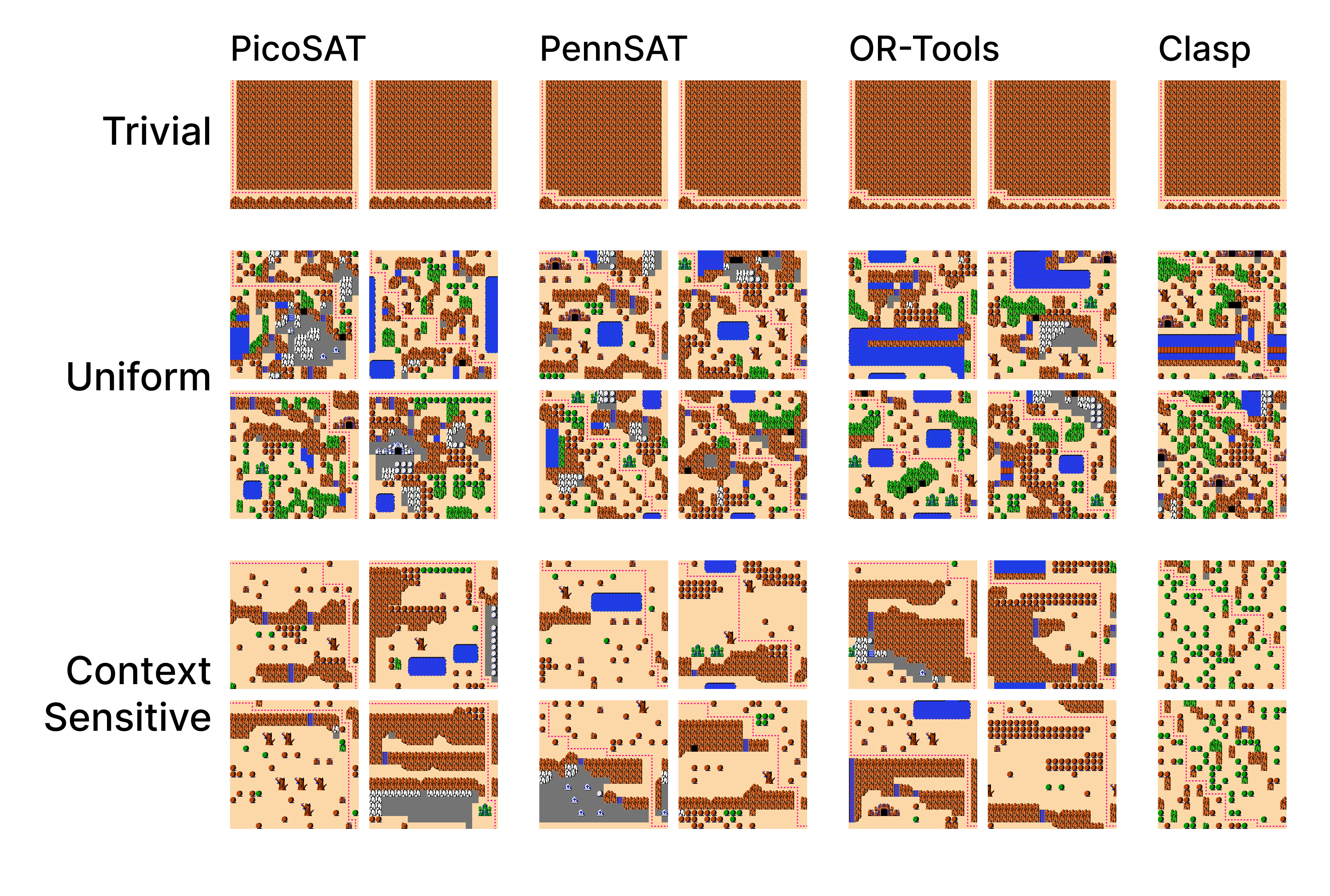}
  \caption{Variation of YORO results within and across different solvers. These \emph{uncurated} samples illustrate the typical range of variation for a given generation problem with different variable orderings. We use different seeds across the different solvers, or else their outputs would be similar or identical. Each solver needs a slightly different configuration in order to get it to frequently consult our pre-rolled decision variable ordering. The context-sensitive outputs from Clasp (the solver for the answer-set solving system Clingo) are different; we were unable to configure Clasp to behave like a more basic SAT solver.}
  \label{fig:different-solvers}
  \Description{This is the description.}
\end{figure*}

To demonstrate that the YORO technique is adaptable to different off-the-shelf SAT solvers, we generated outputs for the \emph{Zelda} domain using the following four different solvers, each of which accepts a SAT formula encoded in the standard DIMACS file format~\cite{prestwich2009cnf}:

\begin{enumerate}
    \item \textbf{PicoSAT}, a small solver by Armin Biere based on MiniSAT \cite{picosat}. We call it in Python via the \texttt{pycosat} bindings \cite{pycosat}.
    \item \textbf{OR-Tools}, Google's operations research toolkit, which includes a SAT-based constraint solver called CP-SAT \cite{ortools}.
    \item \textbf{PennSAT}, a simple, pure-Python SAT solver with the ability to entirely disable heuristics. \cite{pennsat}
    \item \textbf{Clasp}, an answer set programming solver included with the ASP system Clingo \cite{clasp}.
\end{enumerate}

The solving time can vary greatly between solvers and even between inputs. As the size of the output and the complexity of the encoding grows, the solving time may scale poorly. For simple inputs, like in the black-and-white experiment, the solver may find a solution in milliseconds. For complex inputs and global constraints, like in the \emph{Zelda} experiment, solving may take anywhere from a couple seconds to multiple minutes, depending on the solver and random seed. It would be misleading to compare these times with Gumin's original WFC implementation because that system is unable to express the non-local good dirt path constraint.

See Figure~\ref{fig:different-solvers} for several uncurated output samples in the \emph{Legend of Zelda} domain for each of the solvers.

\subsection{Solver Configuration} \label{sec:solver-config}

Modern SAT solvers use a number of heuristics, preprocessing steps, and other advanced procedures to speed up solving time \cite{preprocessing}. However, these techniques can often cause the solver to select variables in an unpredictable order during solving, which can interfere with the effectiveness of the YORO method. While these advanced techniques are often needed to achieve acceptable solving times on hard search problems (where satisfying solutions are exceedingly rare), the kind of problems that often arise in PCG are comparatively easy. Famously, Gumin's WFC algorithm does not implement a backtracking mechanism because contradictions are sufficiently uncommon that rejection sampling (i.e.\ generate-and-test) is sufficient to achieve good performance \cite{karth}.

Fortunately, most SAT solvers are configurable and allow the user to disable advanced techniques, leaving the solver to fall back to its default variable ordering. However, it should be noted that determining a correct configuration is sometimes non-trivial and may require knowledge of the solver's implementation details. In this section, we describe the configurations used for each solver to achieve our results.

The PennSAT solver (designed for teaching purposes) is simple and includes no dynamic heuristics or advanced preprocessing. It uses a static Jeroslow-Wang heuristic which can be disabled.

While the original PicoSAT solver in C is configurable, the \texttt{pycosat} library only provides a minimal, unconfigurable interface. Therefore, we were not able to disable PicoSAT's default dynamic heuristic, which is based on VSIDS \cite{picosat}. However, we claim that VSIDS often has little influence on the order of variable selection here, and we still manage to generate good results for many inputs. As mentioned in Section \ref{sec:variable-ordering}, VSIDS-like heuristics modify the variable ordering in response to contradictions. However, as we previously mentioned, contradictions are rare in the WFC setting.\footnote{This also indicates that, when contradictions are rare, disabling VSIDS should not cause significant increases in solving time.} For inputs that do cause many contradictions, such as those with complex global constraints, PicoSAT's dynamic heuristic may cause deviations from the YORO variable ordering.

The OR-Tools CP-SAT solver has a multitude of configuration parameters. We override the following ones:
\begin{Verbatim}[commandchars=\\\{\}]
model.AddDecisionStrategy(
    all_variables, 
    CHOOSE_FIRST,      \textcolor{gray}{# select vars in ascending order}
    SELECT_MIN_VALUE   \textcolor{gray}{# assign vars to 0 (False) first}
)
\textcolor{gray}{# follow the decision strategy exactly}
solver.parameters.search_branching = FIXED_SEARCH
\textcolor{gray}{# disable preprocessing steps}
solver.parameters.cp_model_presolve = False
\end{Verbatim}

Finally, the Clasp solver also offers a multitude of configuration parameters, which can be customized via command line options. We provide the options \texttt{-{}-heuristic=None}, which disables the VSIDS heuristic, and \texttt{-{}-sat-prepro=no}, which disables preprocessing steps. However, there may be additional features of Clasp that cause deviation from the YORO ordering for certain inputs. While we were able to generate expected outputs using Clasp for the black-and-white and \textit{Mario} experiments, as Figure \ref{fig:different-solvers} shows, our outputs for the \textit{Zelda} experiment using the context-sensitive heuristic were significantly different from the other solvers' outputs.

\subsection{Boolean Formula Transformation}\label{sec:formula-transformation}

While configuration parameters are the most robust method of forcing solvers to respect the variable ordering, some solvers cannot be configured. In this case, it may still be possible to circumvent unpredictable behavior by applying transformations to the Boolean SAT formula before solving. We present two transformations that produce a new formula that is equivalent to the original, but processed differently by the solver. In particular, these transformations influence the solver's \emph{phase selection}; i.e., whether it assigns selected variables to True or False first.

Typical SAT solvers default to assigning variables to False first. Since YORO relies on the solver assigning variables to True in the provided order, this will result in the solver following the \emph{reverse} YORO order. Some solvers allow the user to configure the phase selection strategy. However, rather than relying on configuration, we can solve this problem with the following transformation.

Before solving, negate all Boolean variables in the formula. Intuitively, each variable now represents its semantic negation. For example, this redefines our tile-assignment variables as \\

\texttt{assign}($x, y, t$) $\iff$ position $(x, y)$ is {\bf not} assigned tile $t$\\


Therefore, when the solver sets $\texttt{assign}(x, y, t) = \text{False}$, it represents assigning tile $t$ to cell $(x, y)$ as desired. Before decoding our satisfying assignment back to an output grid, we should negate all literals once more to restore their original meaning.

The second formula transformation we used is a novel strategy to neutralize the influence of PicoSAT's static heuristic on the phase selection. In addition to VSIDS, PicoSAT uses a variant of Jeroslow-Wang in order to determine whether to assign variables to True or False, rather than always defaulting to False \cite{picosat}. We only used this trick when solving with PicoSAT, since as noted in Section \ref{sec:solver-config}, we had no means of disabling its heuristics. This transformation is applied after the negation transformation described above.

Jeroslow-Wang is a statistical heuristic that assigns each Boolean literal $\ell$ an \emph{activity score}, defined as
\begin{align*}
\texttt{activity}(\ell) = \sum_{\text{clause } c \text { containing } \ell} 2^{-|c|} \tag*{\cite{jeroslow-wang}}
\end{align*}
In PicoSAT, when a variable $v$ is selected, the activity scores of its literals are compared. If $\texttt{activity}(v) > \texttt{activity}(\neg v)$, then $v$ is assigned True; otherwise, it is assigned False.

Therefore, in order to ensure that all variables are assigned False first, we pad the formula with trivial length-2 clauses such that for each variable $v$, $\texttt{activity}(v) \leq \texttt{activity}(\neg v)$. Our key observation is that by adding a clause of the form $(d_i \vee \neg v)$, we can increase $\texttt{activity}(\neg v)$ by $2^{-2} = 0.25$ without adding any new constraints. Here $d_i > n$ is a dummy variable that can always be assigned True ($n$ is the number of variables before the transformation). The following pseudocode demonstrates the procedure:
\begin{lstlisting}[mathescape=true]
for each variable $v$ with activity($v$) > activity($\neg v$):
    diff = activity($v$) - activity($\neg v$)
    num_trivial_clauses_to_add = ceil(diff / 0.25)
    for $i$ = 1, 2, ..., num_trivial_clauses_to_add:
        $d_i$ = $n$ + i
        add_clause$\left(d_i \vee \neg v\right)$
\end{lstlisting}

\section{Conclusion}
In this paper, we have shown how the order of decision variables in the definition of a constraint problem can be meaningfully manipulated to successfully shape the statistics of the first solutions output by various off-the-shelf SAT solvers. In particular, we show that it is possible to concentrate all of the randomness into the generation of a matrix of numbers sampled unconditionally from a uniform distribution. With each new randomization of this matrix, we can renumber an existing constraint problem so that the solver will give an appropriately new output sample. This approach shows how general purpose constraint solvers, with their ability to represent and enforce interesting local and global hard constraints, can begin to respect statistical design considerations as well.

\bibliographystyle{ACM-Reference-Format}
\bibliography{references}


\begin{thebibliography}{28}


\ifx \showCODEN    \undefined \def \showCODEN     #1{\unskip}     \fi
\ifx \showDOI      \undefined \def \showDOI       #1{#1}\fi
\ifx \showISBNx    \undefined \def \showISBNx     #1{\unskip}     \fi
\ifx \showISBNxiii \undefined \def \showISBNxiii  #1{\unskip}     \fi
\ifx \showISSN     \undefined \def \showISSN      #1{\unskip}     \fi
\ifx \showLCCN     \undefined \def \showLCCN      #1{\unskip}     \fi
\ifx \shownote     \undefined \def \shownote      #1{#1}          \fi
\ifx \showarticletitle \undefined \def \showarticletitle #1{#1}   \fi
\ifx \showURL      \undefined \def \showURL       {\relax}        \fi
\providecommand\bibfield[2]{#2}
\providecommand\bibinfo[2]{#2}
\providecommand\natexlab[1]{#1}
\providecommand\showeprint[2][]{arXiv:#2}

\bibitem[Bach et~al\mbox{.}(2017)]%
        {bach2017hinge}
\bibfield{author}{\bibinfo{person}{Stephen~H. Bach}, \bibinfo{person}{Matthias Broecheler}, \bibinfo{person}{Bert Huang}, {and} \bibinfo{person}{Lise Getoor}.} \bibinfo{year}{2017}\natexlab{}.
\newblock \showarticletitle{Hinge-Loss Markov Random Fields and Probabilistic Soft Logic}.
\newblock \bibinfo{journal}{\emph{J. Mach. Learn. Res.}} \bibinfo{volume}{18}, \bibinfo{number}{1} (\bibinfo{date}{jan} \bibinfo{year}{2017}), \bibinfo{pages}{3846–3912}.
\newblock
\showISSN{1532-4435}


\bibitem[Bateni et~al\mbox{.}(2023)]%
        {context-sensitive}
\bibfield{author}{\bibinfo{person}{Bahar Bateni}, \bibinfo{person}{Isaac Karth}, {and} \bibinfo{person}{Adam Smith}.} \bibinfo{year}{2023}\natexlab{}.
\newblock \showarticletitle{Better Resemblance without Bigger Patterns: Making Context-Sensitive Decisions in WFC}. In \bibinfo{booktitle}{\emph{Proceedings of the 18th International Conference on the Foundations of Digital Games}} (Lisbon, Portugal) \emph{(\bibinfo{series}{FDG '23})}. \bibinfo{publisher}{Association for Computing Machinery}, \bibinfo{address}{New York, NY, USA}, Article \bibinfo{articleno}{20}, \bibinfo{numpages}{11}~pages.
\newblock
\showISBNx{9781450398558}
\urldef\tempurl%
\url{https://doi.org/10.1145/3582437.3582441}
\showDOI{\tempurl}


\bibitem[Biere(2008)]%
        {picosat}
\bibfield{author}{\bibinfo{person}{Armin Biere}.} \bibinfo{year}{2008}\natexlab{}.
\newblock \showarticletitle{PicoSAT Essentials}.
\newblock \bibinfo{journal}{\emph{Journal on Satisfiability, Boolean Modeling and Computation (JSAT)}}  \bibinfo{volume}{4} (\bibinfo{year}{2008}), \bibinfo{pages}{75--97}.
\newblock


\bibitem[Biere et~al\mbox{.}(2009)]%
        {preprocessing}
\bibfield{author}{\bibinfo{person}{Armin Biere}, \bibinfo{person}{Matti Järvisalo}, {and} \bibinfo{person}{Benjamin Kiesl}.} \bibinfo{year}{2009}\natexlab{}.
\newblock \showarticletitle{Preprocessing in SAT Solving}.
\newblock In \bibinfo{booktitle}{\emph{Handbook of Satisfiability}}, \bibfield{editor}{\bibinfo{person}{Armin Biere}, \bibinfo{person}{Marijn J.~H. Heule}, \bibinfo{person}{Hans van Maaren}, {and} \bibinfo{person}{Toby Walsh}} (Eds.). \bibinfo{publisher}{IOS Press}, \bibinfo{address}{Amsterdam, The Netherlands}, Chapter~9, \bibinfo{pages}{131--153}.
\newblock
\showISBNx{978-1-58603-929-5}
\newblock
\shownote{\url{https://fmv.jku.at/papers/BiereJarvisaloKiesl-SAT-Handbook-2021-Preprocessing-Chapter-Manuscript.pdf}. Accessed 2023}.


\bibitem[Chakraborty et~al\mbox{.}(2014)]%
        {chakraborty2014distribution}
\bibfield{author}{\bibinfo{person}{Supratik Chakraborty}, \bibinfo{person}{Daniel~J. Fremont}, \bibinfo{person}{Kuldeep~S. Meel}, \bibinfo{person}{Sanjit~A. Seshia}, {and} \bibinfo{person}{Moshe~Y. Vardi}.} \bibinfo{year}{2014}\natexlab{}.
\newblock \showarticletitle{Distribution-Aware Sampling and Weighted Model Counting for SAT}. In \bibinfo{booktitle}{\emph{Proceedings of the Twenty-Eighth AAAI Conference on Artificial Intelligence}} \emph{(\bibinfo{series}{AAAI'14})}. \bibinfo{publisher}{AAAI Press}, \bibinfo{address}{Qu\'{e}bec City, Qu\'{e}bec, Canada}, \bibinfo{pages}{1722–1730}.
\newblock


\bibitem[Cipra(1987)]%
        {ising}
\bibfield{author}{\bibinfo{person}{Barry~A. Cipra}.} \bibinfo{year}{1987}\natexlab{}.
\newblock \showarticletitle{An Introduction to the Ising Model}.
\newblock \bibinfo{journal}{\emph{The American Mathematical Monthly}} \bibinfo{volume}{94}, \bibinfo{number}{10} (\bibinfo{year}{1987}), \bibinfo{pages}{937--959}.
\newblock
\urldef\tempurl%
\url{https://doi.org/10.1080/00029890.1987.12000742}
\showDOI{\tempurl}
\showeprint{https://doi.org/10.1080/00029890.1987.12000742}


\bibitem[Cooper(2022)]%
        {sturgeon-org}
\bibfield{author}{\bibinfo{person}{Seth Cooper}.} \bibinfo{year}{2022}\natexlab{}.
\newblock \showarticletitle{Sturgeon: Tile-Based Procedural Level Generation via Learned and Designed Constraints}.
\newblock \bibinfo{journal}{\emph{Proceedings of the AAAI Conference on Artificial Intelligence and Interactive Digital Entertainment}} \bibinfo{volume}{18}, \bibinfo{number}{1} (\bibinfo{date}{Oct.} \bibinfo{year}{2022}), \bibinfo{pages}{26--36}.
\newblock
\urldef\tempurl%
\url{https://doi.org/10.1609/aiide.v18i1.21944}
\showDOI{\tempurl}


\bibitem[Cooper(2023)]%
        {sturgeon-MKIII}
\bibfield{author}{\bibinfo{person}{Seth Cooper}.} \bibinfo{year}{2023}\natexlab{}.
\newblock \showarticletitle{Sturgeon-MKIII: Simultaneous Level and Example Playthrough Generation via Constraint Satisfaction with Tile Rewrite Rules}. In \bibinfo{booktitle}{\emph{Proceedings of the 18th International Conference on the Foundations of Digital Games}} (Lisbon, Portugal) \emph{(\bibinfo{series}{FDG '23})}. \bibinfo{publisher}{Association for Computing Machinery}, \bibinfo{address}{New York, NY, USA}, Article \bibinfo{articleno}{64}, \bibinfo{numpages}{9}~pages.
\newblock
\showISBNx{9781450398558}
\urldef\tempurl%
\url{https://doi.org/10.1145/3582437.3587205}
\showDOI{\tempurl}


\bibitem[Domingos et~al\mbox{.}(2008)]%
        {markov-logic}
\bibfield{author}{\bibinfo{person}{Pedro Domingos}, \bibinfo{person}{Stanley Kok}, \bibinfo{person}{Daniel Lowd}, \bibinfo{person}{Hoifung Poon}, \bibinfo{person}{Matthew Richardson}, {and} \bibinfo{person}{Parag Singla}.} \bibinfo{year}{2008}\natexlab{}.
\newblock \bibinfo{booktitle}{\emph{Markov Logic}}.
\newblock \bibinfo{publisher}{Springer Berlin Heidelberg}, \bibinfo{address}{Berlin, Heidelberg}, \bibinfo{pages}{92--117}.
\newblock
\showISBNx{978-3-540-78652-8}
\urldef\tempurl%
\url{https://doi.org/10.1007/978-3-540-78652-8_4}
\showDOI{\tempurl}


\bibitem[Efraimidis and Spirakis(2006)]%
        {gumbel-sequence}
\bibfield{author}{\bibinfo{person}{Pavlos~S. Efraimidis} {and} \bibinfo{person}{Paul~G. Spirakis}.} \bibinfo{year}{2006}\natexlab{}.
\newblock \showarticletitle{Weighted random sampling with a reservoir}.
\newblock \bibinfo{journal}{\emph{Inform. Process. Lett.}} \bibinfo{volume}{97}, \bibinfo{number}{5} (\bibinfo{year}{2006}), \bibinfo{pages}{181--185}.
\newblock
\showISSN{0020-0190}
\urldef\tempurl%
\url{https://doi.org/10.1016/j.ipl.2005.11.003}
\showDOI{\tempurl}


\bibitem[Gebser et~al\mbox{.}(2007)]%
        {clasp}
\bibfield{author}{\bibinfo{person}{Martin Gebser}, \bibinfo{person}{Benjamin Kaufmann}, \bibinfo{person}{Andr{\'e} Neumann}, {and} \bibinfo{person}{Torsten Schaub}.} \bibinfo{year}{2007}\natexlab{}.
\newblock \showarticletitle{clasp: A Conflict-Driven Answer Set Solver}. In \bibinfo{booktitle}{\emph{Logic Programming and Nonmonotonic Reasoning}}, \bibfield{editor}{\bibinfo{person}{Chitta Baral}, \bibinfo{person}{Gerhard Brewka}, {and} \bibinfo{person}{John Schlipf}} (Eds.). \bibinfo{publisher}{Springer Berlin Heidelberg}, \bibinfo{address}{Berlin, Heidelberg}, \bibinfo{pages}{260--265}.
\newblock
\showISBNx{978-3-540-72200-7}


\bibitem[Golia et~al\mbox{.}(2021)]%
        {golia2021designing}
\bibfield{author}{\bibinfo{person}{Priyanka Golia}, \bibinfo{person}{Mate Soos}, \bibinfo{person}{Sourav Chakraborty}, {and} \bibinfo{person}{Kuldeep~S. Meel}.} \bibinfo{year}{2021}\natexlab{}.
\newblock \showarticletitle{Designing Samplers is Easy: The Boon of Testers}. In \bibinfo{booktitle}{\emph{2021 Formal Methods in Computer Aided Design (FMCAD)}}. \bibinfo{pages}{222--230}.
\newblock
\urldef\tempurl%
\url{https://doi.org/10.34727/2021/isbn.978-3-85448-046-4_31}
\showDOI{\tempurl}


\bibitem[Gumbel(1954)]%
        {gumbel}
\bibfield{author}{\bibinfo{person}{E.~J. Gumbel}.} \bibinfo{year}{1954}\natexlab{}.
\newblock \bibinfo{booktitle}{\emph{Statistical Theory of Extreme Values and Some Practical Applications: A Series of Lectures}}. Vol.~\bibinfo{volume}{33}.
\newblock \bibinfo{publisher}{US Department of Commerce}.
\newblock


\bibitem[Gumin(2016)]%
        {guminWFC}
\bibfield{author}{\bibinfo{person}{Maxim Gumin}.} \bibinfo{year}{2016}\natexlab{}.
\newblock \bibinfo{title}{Wave Function Collapse Algorithm}.
\newblock
\newblock
\urldef\tempurl%
\url{https://github.com/mxgmn/WaveFunctionCollapse}
\showURL{%
\tempurl}


\bibitem[Guzdial et~al\mbox{.}(2022)]%
        {pcgmlbook}
\bibfield{author}{\bibinfo{person}{Matthew Guzdial}, \bibinfo{person}{Sam Snodgrass}, {and} \bibinfo{person}{Adam~J. Summerville}.} \bibinfo{year}{2022}\natexlab{}.
\newblock \bibinfo{booktitle}{\emph{Introduction}}.
\newblock \bibinfo{publisher}{Springer International Publishing}, \bibinfo{address}{Cham}, \bibinfo{pages}{1--6}.
\newblock
\showISBNx{978-3-031-16719-5}
\urldef\tempurl%
\url{https://doi.org/10.1007/978-3-031-16719-5_1}
\showDOI{\tempurl}


\bibitem[Horswill and Foged(2021)]%
        {horswill-path-constraints}
\bibfield{author}{\bibinfo{person}{Ian Horswill} {and} \bibinfo{person}{Leif Foged}.} \bibinfo{year}{2021}\natexlab{}.
\newblock \showarticletitle{Fast Procedural Level Population with Playability Constraints}.
\newblock \bibinfo{journal}{\emph{Proceedings of the AAAI Conference on Artificial Intelligence and Interactive Digital Entertainment}} \bibinfo{volume}{8}, \bibinfo{number}{1} (\bibinfo{date}{Jun.} \bibinfo{year}{2021}), \bibinfo{pages}{20--25}.
\newblock
\urldef\tempurl%
\url{https://doi.org/10.1609/aiide.v8i1.12511}
\showDOI{\tempurl}


\bibitem[Huijben et~al\mbox{.}(2022)]%
        {huijben2022review}
\bibfield{author}{\bibinfo{person}{Iris~AM Huijben}, \bibinfo{person}{Wouter Kool}, \bibinfo{person}{Max~B Paulus}, {and} \bibinfo{person}{Ruud~JG Van~Sloun}.} \bibinfo{year}{2022}\natexlab{}.
\newblock \showarticletitle{A review of the gumbel-max trick and its extensions for discrete stochasticity in machine learning}.
\newblock \bibinfo{journal}{\emph{IEEE Transactions on Pattern Analysis and Machine Intelligence}} \bibinfo{volume}{45}, \bibinfo{number}{2} (\bibinfo{year}{2022}), \bibinfo{pages}{1353--1371}.
\newblock


\bibitem[Jeroslow and Wang(1990)]%
        {jeroslow-wang}
\bibfield{author}{\bibinfo{person}{Robert~G. Jeroslow} {and} \bibinfo{person}{Jinchang Wang}.} \bibinfo{year}{1990}\natexlab{}.
\newblock \showarticletitle{Solving propositional satisfiability problems}.
\newblock \bibinfo{journal}{\emph{Annals of Mathematics and Artificial Intelligence}} \bibinfo{volume}{1}, \bibinfo{number}{1} (\bibinfo{date}{01 Sep} \bibinfo{year}{1990}), \bibinfo{pages}{167--187}.
\newblock
\showISSN{1573-7470}
\urldef\tempurl%
\url{https://doi.org/10.1007/BF01531077}
\showDOI{\tempurl}


\bibitem[Karth and Smith(2017)]%
        {karth}
\bibfield{author}{\bibinfo{person}{Isaac Karth} {and} \bibinfo{person}{Adam~M. Smith}.} \bibinfo{year}{2017}\natexlab{}.
\newblock \showarticletitle{WaveFunctionCollapse is Constraint Solving in the Wild}. In \bibinfo{booktitle}{\emph{Proceedings of the 12th International Conference on the Foundations of Digital Games}} (Hyannis, Massachusetts) \emph{(\bibinfo{series}{FDG '17})}. \bibinfo{publisher}{Association for Computing Machinery}, \bibinfo{address}{New York, NY, USA}, Article \bibinfo{articleno}{68}, \bibinfo{numpages}{10}~pages.
\newblock
\showISBNx{9781450353199}
\urldef\tempurl%
\url{https://doi.org/10.1145/3102071.3110566}
\showDOI{\tempurl}


\bibitem[Katz(2021)]%
        {pennsat}
\bibfield{author}{\bibinfo{person}{Jediah Katz}.} \bibinfo{year}{2021}\natexlab{}.
\newblock \bibinfo{title}{{UPenn CIS 189: Solving Hard Problems in Practice, Lecture 4.}}
\newblock \bibinfo{howpublished}{\url{https://web.archive.org/web/20211228072631/https://www.cis.upenn.edu/~cis189/files/Lecture4.pdf}}.
\newblock
\newblock
\shownote{Accessed 2023}.


\bibitem[Moskewicz et~al\mbox{.}(2001)]%
        {chaff}
\bibfield{author}{\bibinfo{person}{Matthew~W. Moskewicz}, \bibinfo{person}{Conor~F. Madigan}, \bibinfo{person}{Ying Zhao}, \bibinfo{person}{Lintao Zhang}, {and} \bibinfo{person}{Sharad Malik}.} \bibinfo{year}{2001}\natexlab{}.
\newblock \showarticletitle{Chaff: Engineering an Efficient SAT Solver}. In \bibinfo{booktitle}{\emph{DAC '01: Proceedings of the 38th annual Design Automation Conference}} (Las Vegas, Nevada, USA) \emph{(\bibinfo{series}{DAC '01})}. \bibinfo{publisher}{Association for Computing Machinery}, \bibinfo{address}{New York, NY, USA}, \bibinfo{pages}{530–535}.
\newblock
\showISBNx{1581132972}
\urldef\tempurl%
\url{https://doi.org/10.1145/378239.379017}
\showDOI{\tempurl}


\bibitem[Nelson and Smith(2016)]%
        {pcgbook}
\bibfield{author}{\bibinfo{person}{Mark~J. Nelson} {and} \bibinfo{person}{Adam~M. Smith}.} \bibinfo{year}{2016}\natexlab{}.
\newblock \bibinfo{booktitle}{\emph{ASP with Applications to Mazes and Levels}}.
\newblock \bibinfo{publisher}{Springer International Publishing}, \bibinfo{address}{Cham}, \bibinfo{pages}{143--157}.
\newblock
\showISBNx{978-3-319-42716-4}
\urldef\tempurl%
\url{https://doi.org/10.1007/978-3-319-42716-4_8}
\showDOI{\tempurl}


\bibitem[Nintendo(1986)]%
        {zelda-nes}
\bibfield{author}{\bibinfo{person}{Nintendo}.} \bibinfo{year}{1986}\natexlab{}.
\newblock \bibinfo{title}{The Legend of Zelda}.
\newblock \bibinfo{howpublished}{[Family Computer Disk System]}.
\newblock
\urldef\tempurl%
\url{https://nesmaps.com/maps/Zelda/ZeldaOverworldQ1.html}
\showURL{%
\tempurl}


\bibitem[Nintendo(1988)]%
        {mario-nes}
\bibfield{author}{\bibinfo{person}{Nintendo}.} \bibinfo{year}{1988}\natexlab{}.
\newblock \bibinfo{title}{Super Mario Bros. 3}.
\newblock \bibinfo{howpublished}{[Family Computer Disk System]}.
\newblock
\urldef\tempurl%
\url{https://www.spriters-resource.com/resources/sheets/150/153078.png}
\showURL{%
\tempurl}


\bibitem[Perron and Didier(2023)]%
        {ortools}
\bibfield{author}{\bibinfo{person}{Laurent Perron} {and} \bibinfo{person}{Frédéric Didier}.} \bibinfo{year}{2023}\natexlab{}.
\newblock \bibinfo{booktitle}{\emph{CP-SAT}}.
\newblock Google.
\newblock
\urldef\tempurl%
\url{https://developers.google.com/optimization/cp/cp_solver/}
\showURL{%
\tempurl}


\bibitem[Prestwich(2009)]%
        {prestwich2009cnf}
\bibfield{author}{\bibinfo{person}{Steven~D Prestwich}.} \bibinfo{year}{2009}\natexlab{}.
\newblock \showarticletitle{CNF Encodings.}
\newblock \bibinfo{journal}{\emph{Handbook of satisfiability}}  \bibinfo{volume}{185} (\bibinfo{year}{2009}), \bibinfo{pages}{75--97}.
\newblock


\bibitem[Schnell(2023)]%
        {pycosat}
\bibfield{author}{\bibinfo{person}{Ilan Schnell}.} \bibinfo{year}{2023}\natexlab{}.
\newblock \bibinfo{title}{pycosat}.
\newblock
\newblock
\newblock
\shownote{Accessed 2023. Version 0.6.6. \url{https://pypi.org/project/pycosat/}}.


\bibitem[Smith and Mateas(2011)]%
        {asp4pcg}
\bibfield{author}{\bibinfo{person}{Adam~M Smith} {and} \bibinfo{person}{Michael Mateas}.} \bibinfo{year}{2011}\natexlab{}.
\newblock \showarticletitle{Answer set programming for procedural content generation: A design space approach}.
\newblock \bibinfo{journal}{\emph{IEEE Transactions on Computational Intelligence and AI in Games}} \bibinfo{volume}{3}, \bibinfo{number}{3} (\bibinfo{year}{2011}), \bibinfo{pages}{187--200}.
\newblock


\end{thebibliography}


\end{document}